\documentclass[sigconf]{acmart}

\usepackage{booktabs} 
\usepackage{amssymb}
\usepackage{graphicx}
\usepackage{algorithm}
\usepackage{algorithmic}
\usepackage{multirow}
\usepackage{makecell}
\usepackage{url}

\setcopyright{rightsretained}

\acmConference[MLG Workshop'17]{ACM KDD conference}{August 2017}{Halifax, Nova Scotia, Canada} 
\acmYear{2017}
\copyrightyear{2017}

\begin{document}
\title{Adaptive Candidate Generation for Scalable Edge-discovery Tasks on Data Graphs}

\author{Mayank Kejriwal}
\affiliation{%
  \institution{Information Sciences Institute}
  \streetaddress{USC Viterbi School of Engineering}
  \city{Marina Del Rey} 
  \state{CA} 
  \postcode{90292}
}
\email{kejriwal@isi.edu}

\renewcommand{\shortauthors}{Kejriwal}

\begin{abstract}
Several `edge-discovery' applications over graph-based data models are known to have worst-case quadratic time complexity in the nodes, even if the discovered edges are sparse. One example is the generic link discovery problem between two graphs, which has invited research interest in several communities. Specific versions of this problem include link prediction in social networks, ontology alignment  between metadata-rich RDF data, approximate joins, and entity resolution between instance-rich data. As large datasets continue to proliferate, reducing quadratic complexity to make the task practical is an important research problem. Within the entity resolution community, the problem is commonly referred to as blocking. A particular class of learnable blocking schemes is known as Disjunctive Normal Form (DNF) blocking schemes, and has emerged as state-of-the art for homogeneous (i.e. same-schema) tabular data. Despite the promise of these schemes, a formalism or learning framework has not been developed for them when input data instances are generic, attributed graphs possessing both node and edge heterogeneity. With such a development, the complexity-reducing scope of DNF schemes becomes applicable to a variety of problems, including entity resolution and type alignment between heterogeneous graphs, and link prediction in networks represented as attributed graphs. This paper presents a graph-theoretic formalism for DNF schemes, and investigates their learnability in an optimization framework. We also briefly describe an empirical case study encapsulating some of the principles in this paper. 
\end{abstract}

%
%




\maketitle

\section{Introduction}\label{introduction}

Constrained edge-discovery tasks constitute an important class of problems in communities that rely on graph data models \cite{linkminingsurvey}. Examples include \emph{link discovery} (e.g. entity resolution and class matching) in the Semantic Web \cite{rahmsurvey}, \cite{ontology}, and a variety of \emph{link prediction} tasks in network-oriented communities such as social media, bioinformatics and advertising \cite{lp-proteins}, \cite{lp-survey}, \cite{advertising}. Algorithms attempting to solve such tasks take as input either a single graph or two graphs, and predict a set of edges linking nodes in the graphs. The semantics and constraints of the predicted link depends on the task formulation: when performing entity resolution (ER), for example, the link is expected to have \emph{:sameAs} semantics indicating that the two linked entities refer to the same underlying  entity \cite{rahmsurvey}.

A real-world observation about many edge-discovery tasks is that many interesting links are typically \emph{sparse} in the space of all possible edges, which is quadratic in the number of nodes \cite{lp-sparsity}. 
Due to their quadratic complexity, one-step algorithms that predict a link by performing expensive computations on \emph{each} pair of nodes have gradually been superseded by two-step algorithms, especially in the ER community (Section \ref{relatedwork}). In two-step ER, the first step is typically known as \emph{blocking} \cite{bilenkodnf}. Using an indexing function known as a \emph{blocking scheme}, a blocking algorithm clusters approximately similarly entities into (possibly overlapping) clusters known as blocks. Only entities sharing a block are candidates for further analysis in the second \emph{similarity} step. State-of-the-art similarity algorithms in various communities are now framed in terms of machine learning, typically as binary classification \cite{lp-sl}, \cite{linkminingsurvey}. 

This basic two-step framework can also be extended to generic link discovery tasks. As an example, suppose the task is discovery and prediction of co-authorship links between scientists in a social network. Rather than exhaustively evaluate all pairs of (scientist) nodes, we could first index scientists based on a simple condition: the overlap between the keywords used in their papers. Only nodes with sufficient keyword overlap would undergo more expensive computations.  On highly specialized domains, a domain expert might be required to hand-craft appropriate indexing schemes.

Due to expense of manual expertise, \emph{automatic} discovery of such indexing schemes, also using machine learning, was motivated as a research problem in the previous decade \cite{knoblockdnf}, \cite{bilenkodnf}.
Schemes learned in a \emph{supervised} setting are able to adapt to available \emph{training data}, precluding the need for manual hand-crafting. Like in any machine learning framework, the expressiveness of such a scheme would depend both on the underlying properties of the class of schemes (the `hypothesis' space) as well as the learning algorithm optimizing over this space \cite{bishop}.
   
In this paper, we develop a class of schemes known as Disjunctive Normal Form (DNF) schemes for a generic data model called a \emph{directed, labeled attributed data graph model} (Section \ref{datamodel}). The model is designed to be generic enough that several extant graphs of interest, including RDF data and directed, heterogeneous networks, can be expressed as its instances.
In related work (Section \ref{relatedwork}), we describe how the current theory on DNF schemes limits their use to a specific data model (homogeneous tables) and a constrained problem (record deduplication) \cite{knoblockdnf}, \cite{bilenkodnf}, \cite{kejriwaldnf}. Despite their excellent performance in that setting, DNF schemes were never proposed or developed for generic graph models or for sparse edge-discovery tasks (on these graph models) that are less constrained than homogeneous record deduplication (Section \ref{problem}). In Section \ref{dnfschemes}, we present a constructive formalism for DNF schemes that can be applied on graphs (Section \ref{formalism}), followed by results on the \emph{learnability} of these schemes (Section \ref{learnability}). Specific contributions presented in this paper over current state-of-the-art work in DNF indexing are summarized in Table \ref{contributions}.
In Section \ref{applications}, we use our experience with a recent case study to illustrate the empirical utility of graph-theoretic DNF schemes. Although the implementation itself was ad-hoc, and for a very specific problem (entity resolution on Semantic Web datasets), the underlying theory was subsumed by the formalism in this paper. This is also the case for various other DNF schemes proposed in the literature e.g., for homogeneous tabular data without missing values.

\section{Intuition}\label{intuition}
We use Figure \ref{runningexample} to provide some intuition behind the need for adaptive candidate generation. We are looking to discovery `collaborator' links in a knowledge graph describing artists and artistic creations. Due to space limitations, the graph only shows a fragment from the \emph{Movies} domain, but we can imagine the coverage to extend to other artistic domains like songs and screenplays. It is not uncommon for actors, for example, to serve as directors or producers on other projects. In the music domain, many singers write their own songs.

Note that the overall link discovery problem either requires domain knowledge, or is supervised (requires training data), since a system cannot know what it is looking for otherwise. For arbitrary links, only the latter `data-driven' option is feasible. We assume the supervised link discovery framework in the rest of this work.   

To motivate candidate generation, suppose we already know the link discovery function $f$. Such a function is applied on a pair of nodes, and returns the probability that the link in question exists between the pair. Even so, one would have to apply the link discovery function to every pair of nodes, which is infeasible due to its quadratic complexity. 

In \emph{theory}, for an \emph{arbitrary} link discovery function, quadratic complexity is unavoidable.  In \emph{practice}, as the small example in Figure \ref{runningexample} intuits, it is usually unnecessary for \emph{real-world} link types. We could hypothesize, for example, that a collaborator link (either in the past or the future) is unlikely to exist between two nodes unless they are linked to a common artistic work. However, a link may also exist if the two artists are still active and share a close relative. 

The research question addressed by this paper is, given a set of sufficiently representative training examples, where each (negative) positive example is a pair of (non-) linked nodes, how can we \emph{learn} a function that operates in near-linear time to generate a candidate set of promising node pairs that could be further processed by an expensive link discovery algorithm?  

Note that candidate generation is independent from the mechanism of the link discovery algorithm itself. In recent years, link discovery has witnessed a lot of progress due to the advent of deep neural networks e.g., knowledge graph embedding methods, relying on latent space representations of nodes and edges in the knowledge graph, have become quite powerful \cite{kge1}, \cite{kge2}. These algorithms are expensive to train and evaluate, which further motivates the development of adaptive, scalable methods for `good' candidate generation algorithms that significantly reduce the quadratic space without degrading recall. 


\section{Related Work}\label{relatedwork}
Link prediction and entity resolution (ER) were both recognized as important steps in the overall \emph{link mining} community about a decade ago \cite{linkminingsurvey}. In the Semantic Web community, instance matching \cite{rahmsurvey}, link discovery \cite{limes}, \cite{silk} and class matching \cite{ontology} are specific examples of such sparse edge-discovery tasks. Other applications include protein structure prediction (bioinformatics) \cite{lp-proteins}, click-through rate prediction (advertising) \cite{advertising}, social media and network science \cite{lp-survey}, \cite{lp-location}.

The sparsity of positive edges (equivalently known as the \emph{class imbalance} problem in machine learning \cite{bishop}) is well-known in several communities \cite{lp-sparsity}. Blocking methods for ER have continued to be extensively researched, with more recent research in the Semantic Web focused on \emph{data-driven} approaches \cite{hhis}. This work presents \emph{graph-theoretic} formalism and learnability results for a specific class of blocking schemes called Disjunctive Normal Form (DNF) blocking schemes that have rapidly emerged as state-of-the-art for deduplicating \emph{homogeneous tables}  \cite{knoblockdnf}, \cite{bilenkodnf}. We believe this is due to both their strong theoretical foundations, as well as their recently demonstrated experimental robustness, even with noisy training data \cite{kejriwaldnf}. The formalism in prior work is briefly reviewed in Section \ref{background}.

Blocking, as a preprocessing complexity-reduction step, is not the only avenue for addressing scalability. In networks with no edge labels, or otherwise informative property values, structural features are important for predicting missing links \cite{lp-survey}. In large networks, techniques like matrix factorization \cite{lp-matrix}, stochastic optimization \cite{lp-stochastic} and message passing \cite{advertising} are more important than complexity-reduction techniques. Such techniques are complementary, not competitive, with the DNF schemes proposed herein. For example, one could use the blocking techniques to first reduce the pairwise complexity space, setting 0 for any element pairs in the matrix that were not retrieved by the blocking. This results in a sparser matrix and faster computation. Similarly, where edge weight computation is required in a graph, one would only compute weights for edges in the candidate set of node pairs retrieved by blocking, which yields a much sparser graph and faster analytics.  

We also note that such techniques are orthogonal to the Semantic Web, where edge-labels are given by property URIs. Furthermore, on the Web of Linked Data, the usefulness of labels and property values for link discovery problems is well-known, particularly when the data has loose schema bindings \cite{hhis}. Such datasets are becoming increasingly common \cite{linkeddata}. Case-study results presented in Section \ref{applications} demonstrate that, on real-world graph data, the DNF schemes presented in this work can significantly reduce complexity in data-driven link discovery applications.  

\begin{table*}[!t]
\caption{Contributions in this paper compared to prior work.}
    \begin{tabular}[t]{|p{8.5cm} |p{8.4cm}|}
\hline
    {\bf DNF schemes in prior work} & {\bf DNF schemes in this work}\\ \hline
{Specific to homogeneous tables} & {Proposed for heterogeneous graphs}\\ \hline
{Specific to the deduplication task} & {Applicable to any sparse edge-discovery task where training data is available}  \\ \hline
{Handling missing values not evident} & {Addresses the missing value problem} \\ \hline
{Entities (i.e. tabular records) must necessarily have the same type}& {Nodes can have different (and even multiple) types} \\ \hline
{A single tabular instance assumed as input}& {Proposed for edge-discovery tasks in both one-graph and two-graph scenarios} \\ \hline
{Learning as single-step optimization}& {Learning as multi-step optimization} \\ \hline
   
{No reduction results from extant blocking schemes}& {Attribute Clustering \cite{hhis} shown to be a special case (Theorem \ref{ac})} \\ \hline
{No robustness results}& {Empirically robust to noisy training data} \\ \hline
\end{tabular}
\label{contributions}
\end{table*}
\section{Data Model}\label{datamodel}

The specific data model adopted for this work is a \emph{labeled, directed attributed data graph model}. Let $\Sigma$ denote a finite alphabet (e.g. Unicode characters). We refer to an instance of this graph model as a \emph{data graph} (in the spirit of \cite{SAindex}): 
\begin{definition}\label{datagraph}
{\bf Data graph}A \emph{data graph} $G$ is a labeled, multi-relational graph encoded by the 7-tuple $G = (V, E, l_V, A_V, \Sigma_V,$ $\Sigma_E, \Sigma_A)$, where $V$ is the set of nodes, $E \subseteq V \times V \times \Sigma_E$ is the set of directed, labeled edges, $\Sigma_E \subseteq \Sigma^*$ is a finite set denoted as the \emph{edge vocabulary}, $l_V$ is a function mapping a node $v \in V$ to a label in $\Sigma_V \subseteq \Sigma^*$, $\Sigma_V$ is a finite set denoted as the \emph{node vocabulary}, $A_V$ is a partially ordered mapping known as the \emph{attribute mapping} $A_V: V \rightarrow {\Sigma_A}^*$, and $\Sigma_A \subseteq \Sigma^*$ is a finite \emph{attribute vocabulary}.
\end{definition}

\begin{figure}[t]
\centering
\includegraphics[height=4.3cm, width=8.0cm]{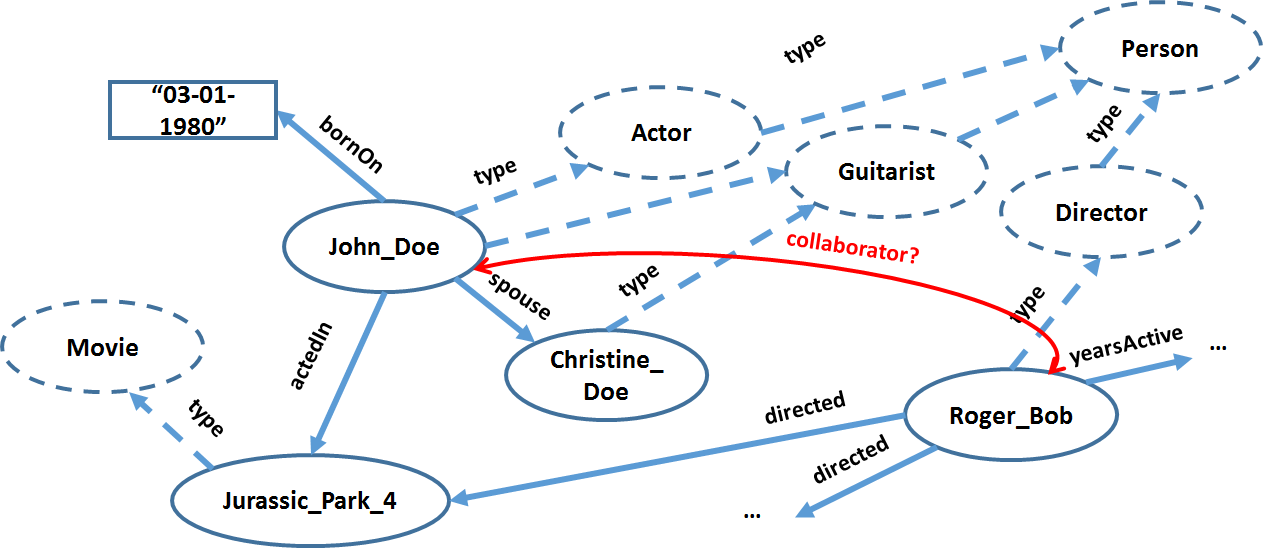}
\caption{Illustration of link discovery, used as the running example throughout this work. Dotted lines (labeled \emph{type}) and nodes respectively represent attributes and attribute mappings (Section \ref{datamodel}).}
\label{runningexample} 
\end{figure}
Per Definition \ref{datagraph}, let $A_V(v)$ represent the \emph{attribute set} of node $v \in V$, and let $l_V(v)$ represent the \emph{label} of node $v$. In slight abuse of notation, we typically refer to a node by its label.
We note that Semantic Web data models like Resource Description Framework (RDF) can be expressed \emph{graph-theoretically} in terms of the defined model, as an RDF graph is just a directed, labeled graph \emph{with constraints}. 
By way of example, one such intuitive constraint is that there does not exist an edge $(v_1, v_2, l) \in E$ such that $l_V(v_1)$ is a \emph{literal}, per some pre-specified predicate (e.g. \emph{isLiteral}) that distinguishes literal elements in $\Sigma_V$ from URI elements. In the same vein, $\Sigma_V$ is constrained so that all non-literal elements (i.e. for which \emph{isLiteral} returns \emph{False}) are necessarily (blank or non-blank) URIs\footnote{$A_V$ returns an empty attribute set for nodes lacking class information. It is \emph{partially ordered} to enable representation of \emph{class hierarchies}.}, and $\Sigma_E$ only contains URIs.

\begin{example}\label{example-dg}
Figure \ref{runningexample} illustrates a data graph that resembles an RDF graph. The dotted \emph{type} edge, which is not formally an edge per Definition \ref{datagraph}, indicates an attribute mapping, with attributes represented as dotted ovals. For example, $A_V$\emph{(John\_Doe)} returns the attribute set \emph{\{Actor, Guitarist\}}. Note also that, per RDF convention, we have placed the literal ``03-01-1980" in a rectangle; Definition \ref{datagraph} does not actually distinguish between literal and non-literal nodes.  
\end{example}
 
In the context of RDF/OWL\footnote{Web Ontology Language \cite{owl}} data, Example \ref{example-dg} illustrates that attributes in Definition \ref{datagraph} typically serve the same role as a set of \emph{ontological classes}. Since class hierarchies (i.e. super-classes and sub-classes) are prevalent in expressive ontologies, a node is permitted multiple attributes (i.e. an attribute \emph{set}). In the model proposed in \cite{SAindex}, only a single attribute per node was permitted, and edges were necessarily undirected and unlabeled. Such graphs are special cases of Definition \ref{datagraph}, and the findings in that paper\footnote{Namely, building efficient index data structures for speedy query processing.} are complementary to the formalism presented herein. Note also that all three vocabularies in Definition \ref{datagraph} may be \emph{empty}. Essentially, nodes and edges are allowed to be unlabeled and untyped.

Whenever two graphs are indicated, subscripts on the relevant notation will be used to make a distinction. A specific caveat is the usage of the term \emph{attribute}. In graph-theoretic terminology, adopted herein for the sake of generality, \emph{nodes} are attributed \cite{SAindex}, meaning that an attribute is like an RDF class. This is in contrast to \cite{hhis}, where an \emph{attribute} was a set of pairs, with each pair consisting of an edge-label (i.e. an RDF property URI) and an object value. Finally, note that the \emph{finiteness} of the various elements in $G$ is motivated primarily by real-world applications on finite data graphs. Technically, permitting the sets to be \emph{countably infinite} does not fundamentally alter the subsequent formalism, but does make it unnecessarily more involved.

\section{Problem Formulation}\label{problem}

With the data model in place, there are two problem scenarios within the scope of this paper. The first, denoted as the \emph{one-graph} scenario, concerns sparse edge-discovery tasks on a single data graph input. Given a data graph $G$, let there be an \emph{unknown} partition of the quadratic space $V \times V$ into two sets $P$ (links) and $N$ (non-links). We assume a sparsity condition i.e. $|N| = \rho(|N|+|P|)$ ($\rho \approx 1.0$, but is strictly less than 1.0), and an available training set (sampled i.i.d) $T=P_T \cup N_T$, where $P_T \subset P$  and $N_T \subset N$. We denote $\rho$ as the \emph{optimal reduction ratio (RR)}. The \emph{pairwise complexity-reduction problem} is to learn a sufficiently expressive scheme that, when executed on $G$, results in a candidate set $C$ of node-pairs such that the \emph{empirical} RR ($1.0-|C|/|V|^2$) is maximized while ensuring that the positive link coverage (the \emph{Pairs Completeness} or PC), defined as $|C \cap P|/|P|$, is above a minimum pre-specified threshold in \emph{expectation}. This learning problem is formally expressed as an optimization program in Section \ref{learnability}. 

Concerning the \emph{two-graph} scenario, the problem is similarly defined as above, except that all links must be in the set $V_1 \times V_2$, with $V_1, V_2$ being the node sets of the two data graph inputs $G_1, G_2$ respectively. Rather than adopt separate formalisms in Section \ref{formalism} for the two scenarios, we frame the definitions, where relevant, in a way such that (1) two data graphs are never required to be distinct and can therefore be the same graph, and (2) two nodes    are always required to be distinct. By maintaining (1) and (2) throughout the construction, the one-graph and two-graph treatments are unified, unlike in prior work on the subject \cite{bilenkodnf}, \cite{knoblockdnf}, \cite{hhis}. Theoretically, self-link discovery is also avoided.

Most importantly, the complexity-reduction problem studied in this paper is \emph{agnostic} to the underlying link specification function (LSF), since the ground-truth partitioning of the quadratic node-pair space (into links and non-links) is unknown (and can be arbitrary). This is in contrast to existing complexity-reduction systems in the literature wherein either the LSF itself or its semantics, is \emph{known} \cite{limes}, \cite{silk}, \cite{hhis}.  

\section{Disjunctive Normal Form Schemes}\label{dnfschemes}
\subsection{Background: DNF blocking schemes for tabular deduplication}\label{background}
\begin{figure}[t]
\centering
\includegraphics[height=4.2cm, width=8.0cm]{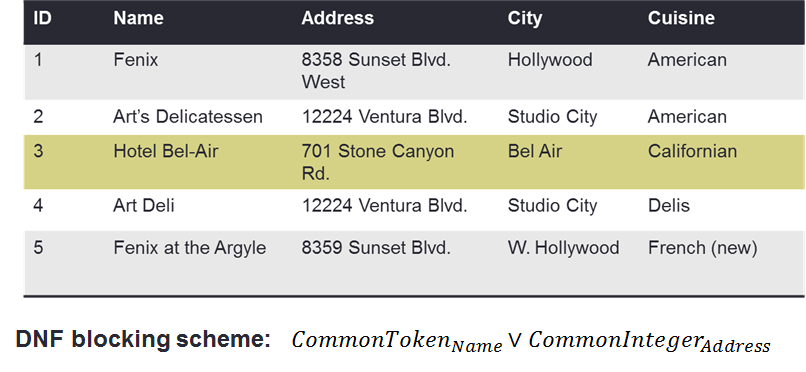}
\caption{Example of a DNF blocking scheme for the deduplication task on a tabular \emph{Restaurants} benchmark. The scheme takes as input a pair of tabular records and returns \emph{True} (otherwise \emph{False}) if they should be included in the candidate set $C$ (Section \ref{problem}).}
\label{dedup-dnf} 
\end{figure}

The theory (i.e. formalism and learnability) for an adaptive class of complexity-reduction schemes, called \emph{Disjunctive Normal Form (DNF) blocking schemes}, is especially well-developed for the task of \emph{homogeneous tabular deduplication} \cite{knoblockdnf}, \cite{bilenkodnf}. Figure \ref{dedup-dnf} illustrates such a scheme by way of an example. The scheme is given by a Boolean DNF expression that can be arbitrarily complex\footnote{With a finite set of $n$ predicates (e.g. $CommonToken_{Name}$ in Figure \ref{dedup-dnf}), there are $2^{2^n}$ canonical (i.e. arrangement-insensitive) \emph{positive} DNF formulae. Negated literals are not allowed in blocking constructions.} in principle, although in practice, the complexity of the scheme is curbed by a specified parameter. As the mnemonic notation suggests, the scheme takes a pair of entities (tabular records in this case) as input and returns \emph{True} if they share a token in their \emph{Name} column or an  integer in their \emph{Address} column. The predicates that comprise the atoms in the DNF expression are compositions of a function (e.g. \emph{CommonToken}) and a column (e.g. \emph{Name}). Given a set of $g$ such functions (known as \emph{general predicates} \cite{bilenkodnf}) and a table with $c$ columns, a legal DNF scheme is expressible over $gc$ atoms. Given training sets of duplicates and non-duplicates, learning a scheme can be framed in terms of solving an optimization problem over the training sets (Section \ref{learnability}) \cite{bilenkodnf}, \cite{knoblockdnf}.

\subsection{Constructive formalism for sparse edge-discovery on data graphs}\label{formalism}

The basic treatment of DNF blocking schemes in Section \ref{background} illustrates that, at the highest level, there are two crucial components to their construction. The first is akin to the \emph{feature design} phase (of typical machine learning), and corresponds to the choice of predicate functions (e.g. \emph{CommonToken} in Figure \ref{dedup-dnf}). The second is the learning algorithm itself (i.e. choosing and combining the atomic predicates into a complete DNF expression), akin to the \emph{parameter estimation} (e.g. by applying statistical inference techniques on available training data) phase \cite{bishop}.  

In the graph model, these two components are not, in themselves, adequate because of the presence of both node and edge heterogeneity. Node heterogeneity arises because nodes may have different sets of attributes associated with them, while edge heterogeneity arises because of edge labels. Different entities may have different sets of `properties' or edge labels associated with them. A naive adoption of the treatment in Section \ref{background} to the graph-theoretic case runs into the \emph{missing value} problem\footnote{This becomes apparent if each column in the table in Figure \ref{dedup-dnf} is thought of as a `property' or edge label. Every entity is constrained to possess this exact set of properties (the table schema) in the homogeneous tabular deduplication task.}.  

To accommodate heterogeneity and missing values at the conceptual level, additional technical machinery is needed. In the rest of this section, we `construct' the formalism by defining some of these concepts and illustrating them using the running example in Figure \ref{runningexample}. In keeping with practical constraints and intuitions, we impose finiteness constraints on the relevant definitions.  

As in the rest of this paper, we assume an alphabet $\Sigma$. Given a data graph $G=(V, E, l_V, A_V, \Sigma_V, \Sigma_E, \Sigma_A)$, recall that each of $\Sigma_V, \Sigma_E$ and $\Sigma_A$ is a subset of $\Sigma^*$. Using standard terminology from formal automata theory, an arbitrary element from $\Sigma^*$ is referred to as a \emph{string}. An arbitrary element from $\Sigma_V, \Sigma_E$ and $\Sigma_A$ is referred to as a \emph{node label}, \emph{edge label} and \emph{attribute} respectively. 

With these assumptions, we start by defining shallow and deep extractors, which are the most basic (`primitive') units in constructing a DNF scheme:

\begin{definition}\label{pse}
{\bf Primitive shallow extractor (PSE)} Given a graph $G$ and an alphabet $\Sigma$, a \emph{primitive shallow extractor} (PSE) $P_s: \Sigma_L \rightarrow 2^{\Sigma^*}$ is defined as a mapping that takes a node label from $\Sigma_L$ as input and returns (`extracts') a finite set of strings ($\subset \Sigma^*$) as output.
\end{definition}

\begin{example}\label{example-pse}
An example of a PSE would be tokenizing a string into a set (of tokens) based on standard delimiters. For example, using the delimiter - on the date literal ``03-01-1980" in Figure \ref{runningexample}, a set \{``03", ``01", ``1980"\} is obtained. A useful practice is to represent such extractors mnemonically (e.g. \emph{TokenizeString}).
\end{example}

\begin{definition}\label{pde}
{\bf Primitive deep extractor (PDE)} Given a graph $G$ and an alphabet $\Sigma$, a \emph{primitive deep extractor} (PDE) $P_d: 2^{\Sigma^*} \rightarrow 2^{\Sigma^*}$ is defined as a mapping that takes a finite set of strings as input and returns (`extracts') a finite set of strings as output.
\end{definition}

\begin{example}\label{example-pde}
Continuing from Example \ref{example-pse}, an example of a PDE would be \emph{AddOneToIntegers}. It takes a set as input, and for every integer in the set, parses and increments the integer and adds it back to the set (designed for more robust performance against noisy integer inputs \cite{bilenkodnf}). On the input set \{``03", ``01", ``1980"\}, the output would be \{``03", ``01", ``1980", ``04", ``02", ``1981"\}. Another example, designed for text, is to remove stop-words (e.g. \emph{the}) from the set\footnote{Thus, the output set can potentially be smaller (even empty) than the input set.}.
\end{example}

The examples above indicate that the PSEs and PDEs must necessarily be specified by the user. Typically, this is not a bottleneck; authors in several communities have already proposed a wide variety of practical functional classes (e.g. phonetic, token-based, set-based and numeric) \cite{bilenkodnf}, \cite{kejriwal-full}, \cite{lp-survey}. Henceforth, we assume the availability of finite sets $\mathcal{P}_s$ and $\mathcal{P}_d$ of PSEs and PDEs respectively.

\begin{definition}\label{feo}
{\bf Feature extraction operator (FEO)} Given a graph $G$, and extractor sets $\mathcal{P}_s$ and $\mathcal{P}_d$, a \emph{feature extraction operator} (FEO) is a mapping that takes a node $v \in V$ as input, computes its label $l_V(v)$, and performs a finite, non-empty sequence of extraction operations to output a set of strings.
\end{definition}

Given an FEO parameterized by $n$ extractors, it is necessarily the case (per Definitions \ref{pse} and \ref{pde}) that the first extraction, which always exists per Definition \ref{feo}, is \emph{shallow}, and the following $n-1$ extractions (if $n >1$) are \emph{deep}. 

\begin{example}\label{example-feo}
Consider a free-text literal ``Died on 03-03-1943". A first step, as discussed in Example \ref{pse}, is to derive a set of tokens from the literal. Next, as discussed in Example \ref{pde}, the integers could be supplemented with increments\footnote{If a token cannot be parsed as an integer, we design the PDE \emph{AddOneToIntegers} to ignore it.}, but also the \emph{stop-word} `on' should be removed, and the word `died' should be \emph{stemmed} to its canonical form `die'. Functionally, this FEO is represented by the composite mapping \emph{StemWords(RemoveStopWords(AddOneToIntegers(\\TokenizeString($l_V(.)$))))}. 
\end{example}

One issue with the definition of an FEO is that it only operates on the label of the node. In RDF graphs, in particular, the label does not contain enough discriminative information\footnote{In cases such as Freebase, the `label' as defined here is usually an \emph{opaque} URI representing the subject of the entity.}. It becomes necessary to seek out information that is one or more edges (i.e. a \emph{trail}\footnote{The data graph, as defined in Section \ref{datamodel}, is not required to be \emph{acyclic}. This is why, in the subsequent formalism, we refer to trails (which may have cyclical subsequences) and not \emph{paths}. For practical purposes, this subtlety applies more to networks, where cycles are common, than to RDF graphs.}) away. Given a graph $G$, a node $v \in V$ and a sequence $s$ of $n$ edge labels, let a trail $t$, defined as an alternating sequence of nodes and edges in $G$, be denoted as being \emph{valid} if (1) the starting node in $t$ is $v$, (2) the subsequence of edges in the trail corresponds exactly to $s$. 

\begin{example}\label{validpath}
In Figure \ref{runningexample}, the starting node \emph{John\_Doe} and edge-label sequence \emph{(actedIn)} yields a valid trail: \emph{(John\_Doe, actedIn, Jurassic\_Park\_4)}. If John Doe acted in multiple movies, there would be multiple valid trails.
\end{example}

In general, given an edge-label sequence and a starting node, a (possibly empty) set of valid trails can be constructed. Let the terminating node in a trail $t$ be denoted by the symbol $last(t)$. In a slight abuse of notation, let the set of \emph{all} edge-label sequences of length exactly $n$ be denoted by the symbol $\Sigma_E^n$. Similarly, let $T^n$ denote the set of all trails with exactly $n$ edges\footnote{These symbols assume a graph $G$. For more than one graph, subscripts will be used to make a distinction.}. Using these symbols, let $trails(v, s)$ represent the mapping that takes a starting node $v \in V$ and an edge-label sequence $s \in \Sigma_E^n$ as input, and returns a (possibly empty) set of valid trails $T^n \subset T^*$, where $T^*$ is the (\emph{countably infinite}, in the general case) set of all possible trails in graph $G$. 

\begin{definition}\label{t-feo}
{\bf Trail-sensitive feature extraction operator (t-FEO)} Given a graph $G$ and an FEO $f$, a \emph{trail-sensitive feature extraction operator (t-FEO)} is a mapping that takes a node $v \in V$ and a finite sequence $s \in \Sigma_E^n$ with exactly $n \geq 0$ edge labels as input, and for $n=0$, returns $f(v)$. For $n>0$, the operator constructs the set $T^n=trails(v, s)$ and returns (1) the empty set if $T^n$ is empty, (2) $\bigcup_{t \in T^n} f(last(t))$ if $T^n$ is non-empty.
\end{definition}

Notationally, we denote a t-FEO as being \emph{parameterized}\footnote{Given a graph $G$ (\emph{context}) and a non-negative integer $n$ (\emph{hyperparameter}), the t-FEOs represent a \emph{class} of mappings with one degree of freedom (the \emph{parameter} $f$).} by FEO $f$, and with a node $v$ and finite edge-label sequence $s$ as \emph{inputs}. 

\begin{example}\label{example-t-feo}
Consider a t-FEO parameterized by the FEO $f$ defined in Example \ref{example-feo} on the data graph in Figure \ref{runningexample}.  Given the node \emph{John\_Doe} and the simple unit-length edge-label sequence \emph{(bornOn)} as inputs, the t-FEO returns the same output as in Example \ref{example-feo}. On the input \emph{Christine\_Doe} and the same sequence, the t-FEO returns \{\}.
\end{example}

Henceforth, we assume a finite set $F^{\leq n}$ of t-FEOs (with hyperparameters that do not exceed $n$), which can be constructed by bounding $n$ and using a finite set of FEOs. Definition \ref{t-feo} gracefully handles missing values by returning the empty set when the set $T$ of valid trails is also empty. Furthermore, allowing an edge-label sequence to be empty ($n=0$) enables an FEO in Definition \ref{feo} to be cast as a \emph{special case} of a trail-sensitive FEO in Definition \ref{t-feo}.

A t-FEO always operates on a \emph{single} node, while edge-discovery is a pairwise operation. Given two (not necessarily distinct)  t-FEOs from  two \emph{distinct} nodes, either from a single graph (one-graph scenario) or two different graphs (two-graph scenario),  parameterized t-FEOs can be applied on the respective nodes to obtain two \emph{feature-sets} $Z_1$ and $Z_2$. 

A \emph{set-based relation} can now be used to derive a Boolean value from these two sets. Such a relation takes the two sets as inputs and maps them to \emph{True} or \emph{False} based on some condition. While any condition can be used, in theory, the motivation behind developing DNF schemes is to avoid quadratic comparisons, and the relation must be amenable to efficient execution. A specific example of such a relation is the \emph{thresholded Jaccard}, defined as the condition $|Z_1 \cap Z_2|/|Z_1 \cup Z_2| > \theta$, where $\theta$ is a specified threshold. An important, highly effective case in the blocking community is $\theta = 0$, as checking for a single element common to the sets becomes sufficient (and \emph{inverted indexing} techniques become applicable). The rest of this section assumes this simple case; the case of arbitrary real-valued thresholds is left for future work.

Using a set-based relation $R$, and the definitions thus far, a \emph{trail-sensitive predicate} is defined below. Such predicates eventually serve as the atoms in the final DNF construction (similar to the role served by $CommonToken_{Name}$ in Figure \ref{dedup-dnf}).

\begin{definition}\label{t-P}
{\bf Trail-sensitive predicate (t-P)} Given a set-based relation $R$, two t-FEOs $f_1$ and $f_2$ and two finite sets $S_1 \subset \Sigma_{E_1}^*$ and $S_2 \subset \Sigma_{E_2}^*$ of edge-label sequences, defined respectively on two graphs $G_1$ and $G_2$, a \emph{trail-sensitive predicate} (t-P) is a binary relation parameterized as a 5-tuple $(R, f_1, f_2, S_1, S_2)$. A t-P takes as input two distinct nodes $v_1 \in V_1$ and $v_2 \in V_2$, computes the set $Z_1=\bigcup_{s \in S_1}f_1(v_1, s)$ (and similarly, set $Z_2$), and returns $R(Z_1, Z_2)$.
\end{definition}

By bounding any sequence in the edge-label sequence sets $S_1$ and $S_2$ in the definition above\footnote{This requirement is less restrictive than it seems, since every data graph is assumed to have finite \emph{diameter}, which can serve as the theoretical bound, and $\Sigma_E$ was declared finite (Definition \ref{datagraph}).}, the set of all trail-sensitive predicates (denoted as the \emph{predicate universe} $U$) is also finite\footnote{$R$ is presently fixed, and both t-FEOs are necessarily drawn from finite sets per Definition \ref{t-feo} and the note following it.}. Intuitively, these predicates serve as \emph{atoms}, which can now be used to construct general DNF expressions.

One issue is that, so far, the \emph{attributes} of the nodes involved (i.e. node heterogeneities) have been neglected. This issue is addressed by defining an \emph{attribution relation} below:

\begin{definition}\label{ar}
{\bf Attribution Relation} Given two (not necessarily distinct) graphs $G_1$ and $G_2$, an \emph{attribution relation} is a binary relation defined on the attribute mappings $A_{V_1}$ and $A_{V_2}$.  Functionally, it takes as input two distinct nodes $v_1 \in V_1$ and $v_2 \in V_2$, and returns \emph{True} iff some attribute pair in $A_{V_1}(v_1) \times A_{V_2}(v_2)$ is in the relation, and returns \emph{False} otherwise.
\end{definition}

\begin{example}\label{example-ar}
A good (in a \emph{data-driven} sense) attribution relation for the example in Figure \ref{runningexample} is \emph{\{(Actor, Director), (Guitarist, Guitarist)\}}. A safer (in a coverage sense) but more coarse-grained (i.e. less discriminative) relation is \emph{\{(Person, Person)\}}. Note that, for either relation, including \emph{(Movie, Person)} in the relation is inappropriate, since discovering only \emph{collaborator} links is of interest. 
\end{example}

Technically, discovering an appropriate attribution relation is within the scope of the multi-step optimization problem outlined in Section \ref{learnability}. In practice (for reasons described in that section), the problem is constrained enough for an inexpensive external algorithm (e.g. ontology matching) to be used instead \cite{ontology}.

\begin{definition}\label{adnf}
{\bf Attribute-aware DNF scheme} Given a predicate universe $U$, two (not necessarily distinct) graphs $G_1$ and $G_2$ and an attribution relation $\mathcal{A}$, an \emph{attribute-aware Disjunctive Normal Form (DNF) scheme} $D_{\mathcal{A}}$ is a positive\footnote{That is, negated atoms from $U$ are not permitted in the construction.} DNF expression $D$ composed of the atoms in $U$. It takes as input two distinct nodes $v_1 \in V_1$ and $v_2 \in V_2$, and returns True iff $D$ is \emph{True} and either $\mathcal{A}$ is empty or $\mathcal{A}(v_1, v_2)$ is \emph{True}, and returns \emph{False} otherwise.
\end{definition}

\begin{definition}\label{cdnf}
{\bf Composite DNF scheme} A \emph{Composite DNF scheme} $\mathcal{C}$ is defined as a finite set  of attribute-aware DNF schemes that takes as input two distinct nodes $v_1 \in V_1$ and $v_2 \in V_2$, and returns \emph{True} (otherwise \emph{False}) iff there exists a scheme $D_{\mathcal{A}} \in \mathcal{C}$ that returns \emph{True} for the pair $(v_1, v_2)$.
\end{definition}

\begin{example}\label{example-dnf}
Assuming two attribution relations \emph{\{(Actor, Director)\} and \{(Guitarist, Guitarist)\}}, an attribute-aware DNF scheme could be devised for each of the two relations. If the training data is representative, the two schemes would presumably be different. The composite scheme may be thought of as a `committee' of these two schemes. Given two distinct nodes as input, it returns \emph{True} iff either one of the attribute-aware schemes returns \emph{True}, and the corresponding attribution relation is satisfied.
\end{example}

Concerning \emph{execution} of a given DNF blocking scheme on two (not necessarily distinct) graphs $G_1$ and $G_2$ to derive a highly reduced candidate set of node-pairs (recall the original problem in Section \ref{problem}), it can be shown that, under practical constraints (e.g. finiteness and boundedness), a near linear-time indexing algorithm can be applied on the graphs using a given scheme. In the Semantic Web, an example of one such algorithm is \emph{block purging} \cite{hhis}. 

\subsection{Learnability}\label{learnability}

Section \ref{formalism} presented a formalism for constructing (composite) DNF schemes on entire graphs. Given graph inputs, and training sets $P_T$ and $N_T$ of links and non-links, we would ideally like to \emph{learn} a DNF scheme from the training data. This section formally explores the learnability of \emph{unconstrained} DNF schemes. 

As with many learning problems, learning a composite DNF scheme can be framed in terms of solving an optimization problem. We assume as inputs two (not necessarily distinct) graphs $G_1=(V_1, E_1, l_{V_1}, A_{V_1}, \Sigma_{V_1}, \Sigma_{E_1}, \Sigma_{A_1})$ and $G_2=(V_2, E_2, l_{V_2}, A_{V_2}, \Sigma_{V_2}, \Sigma_{E_2}, \Sigma_{A_2})$, training sets $P_T$ and $N_T$ and a finite predicate universe $U$. Let $\mathbf{C}$, denoted as the \emph{hypothesis space}, be the set of all composite DNF schemes that can be legally composed on graphs $G_1$ and $G_2$, using the predicate universe $U$. The optimization problem is stated as:
\begin{equation}
argmin_{\mathcal{C} \in \mathbf{C}} |\{ (v_1, v_2) \in N_T | \mathcal{C}(v_1, v_2) \}|
\end{equation}
subject to the condition that,
\begin{equation}
|\{ (v_1, v_2) \in P_T| \mathcal{C}(v_1, v_2) \}| \geq \epsilon|P_T|
\end{equation}

We denote $\epsilon$ as the \emph{minimum Expected Pairs Completeness (mEPC)}. Intuitively, the optimization program states that an `optimal' composite scheme minimizes the number of negative training examples (the non-links) covered (Eqn. 1), while exceeding a required level of recall (i.e. $\epsilon$) with respect to the positive examples (links), at least in expectation\footnote{The empirical PC of any scheme $\mathcal{C}$ on a given training set is, in fact, the expected PC relative to a full ground-truth, since the training set is sampled i.i.d (Section \ref{problem}).}. Note that, like other optimization problems, the problem above can be stated as a \emph{decision} problem, by asking if a composite scheme exists, such that the fraction of negative examples covered does not exceed (a specified parameter) $\eta$.

The composite scheme $\mathcal{C}$ is necessarily a \emph{finite} set by virtue of $U, \Sigma_{A_1}$ and $\Sigma_{A_2}$ being finite.  Intuitively, any solution to Eqns. (1)-(2) may be thought of as a multi-step procedure. First, the attribution relations governing the scope of each attribute-aware DNF scheme in the composite scheme need to be determined. Next, for each such relation, an attribute-aware DNF scheme needs to be learned. In the worst case, the two steps would not be independent: choosing the wrong relations could result in a sub-optimal composite scheme, even if each individual attribute-aware DNF scheme is optimal with respect to the training examples `covered' by its corresponding attribution relation.  

Given this dependency and the expressiveness of DNF schemes, a natural question is if a tractable solver for Eqns. (1)-(2) exists. The following theorem provides strong evidence against such an existence.
\begin{theorem}\label{npc}
The decision version of  Eqns. (1)-(2) is NP-hard. 
\end{theorem}     
\emph{Proof Intuition:} In prior work on DNF blocking scheme learning for homogeneous tabular deduplication \cite{bilenkodnf}, a simpler version of the decision problem was shown to be NP-hard, by demonstrating a reduction from a known NP-hard problem. In an extended report\footnote{Accessed at the author's arXiv page: \url{https://arxiv.org/pdf/1605.00686.pdf}}, we demonstrate a similar reduction.

Theorem \ref{npc} illustrates a natural tradeoff between the \emph{expressiveness} of DNF schemes (when they are not subject to any constraints) and their learning properties. Generally, edge-discovery tasks are rarely unconstrained. For example, if the task is entity resolution in the Semantic Web, a first step is to use ontology alignment to bound the possible attribute relations \cite{ontology}. In the next step, an approximate attribute-aware DNF scheme learning (for each attribute relation output by the ontology aligner) can be learned. In prior work on DNF schemes, a variety of greedy approximation algorithms have already been proposed for the homogeneous tabular deduplication task, including beam search \cite{knoblockdnf}, greedy set covering \cite{bilenkodnf}, and feature selection \cite{kejriwaldnf}. In recent work, we developed and evaluated  an approximation algorithm for entity resolution on RDF graphs \cite{kejriwal-full}. The empirical results are discussed in Section \ref{applications}.  

\section{Case Study}\label{applications}
Although the primary developments in this work were theoretical, they were motivated by practical large-scale issues in graph-based ecosystems such as the Semantic Web. Recently, we designed an unsupervised entity resolution (ER) system for \emph{schema-free} (i.e. heterogeneous) RDF data \cite{kejriwal-full}. Using bounded parameters and a set of 28 manually crafted extractors (Definitions \ref{pse} and \ref{pde}), we presented an approximation algorithm to learn DNF schemes from training data. Note that, because the system was designed to be unsupervised, a heuristics-based component called a \emph{training set generator} (TSG) was also a part of the system; the training examples used to bootstrap the learning processes in the entire system were output by this TSG. Due to its unsupervised nature, the TSG could make mistakes: in many cases, the precision of the generated training set was well below 80\%. This, in turn, imposed a strong \emph{robustness} requirement on the entire system, especially blocking scheme learning \cite{kejriwal-full}.  

To evaluate DNF blocking scheme learning, we gathered a set of ten RDF test cases\footnote{These test cases are detailed in the original journal article where we described the overall unsupervised entity resolution system \cite{kejriwal-full}.}, and used a token-based blocking algorithm known as \emph{token-based Attribute Clustering}\footnote{As indicated at the end of Section \ref{datamodel}, an `attribute' in \cite{hhis} was defined as a set of \emph{edge label-object value} pairs associated with an entity (a node in the data graph). Herein, the word was used in the traditional graph-theoretic sense.} (AC) as a baseline \cite{hhis}. The AC algorithm was designed for the two-graph scenario mentioned in Section \ref{problem}. It is \emph{non-adaptive}; the algorithm uses a pre-defined similarity model to cluster edge-label sets  $\Sigma_{E_1}$ and $\Sigma_{E_2}$. An example of a similarity model is using instance-based measures (like cosine similarity) on corresponding object-values. Once the clusters are obtained, entities can be assigned to blocks based on whether they share common tokens (or by extension, other features) in at least two object values corresponding to edges that were assigned to the same cluster \cite{hhis}.

Experimental results were reported in \cite{kejriwal-full}, with the metrics being Pairs Completeness (PC), Reduction Ratio (RR) and their harmonic mean (F-score). PC and RR were earlier defined in Section \ref{problem}. The results in \cite{kejriwal-full} showed that, due to its adaptive nature, the DNF approximation algorithm (1) was able to outperform AC on the F-score metric on six out of ten test cases, (2) achieved a mean RR that was over 7.5\% higher than that achieved by AC, with a mean loss in PC just below 2.6\%, and (3) had stable RR performance, with 2.42\% standard deviation across all ten test cases, compared with 13.13\% deviation for AC . (3), in particular, shows that adaptive DNF learning is \emph{reliable} compared to AC, which can be an important concern in real-world linkage scenarios that exhibit \emph{dynamicity}, such as Linked Open Data \cite{linkeddata}. Even with noisy training data, the learner continued to exhibit stable RR\footnote{Importantly, high, stable RR is essential for \emph{high volume} tasks because RR grows \emph{quadratically} with the number of nodes, and even small  improvements or variations (less than a percent) disproportionately impact candidate set size.}. The competitive performance with AC shows that the DNF schemes are applicable to \emph{schema-free} data.

Concerning the run-time of the blocking itself, both blocking methods above had similar run-times, which were much smaller than the full edge discovery problem (involving feature computation as well as application of a machine learning-based similarity function).

Post-hoc error analyses suggested at least two possible reasons behind the case study performance numbers reported in \cite{kejriwal-full}. First, it could be the case that AC is not as \emph{expressive} as DNF blocking. The following theorem formalizes this intuition, with a proof provided in the extended report: 

\begin{theorem}\label{ac}
There exists a finite predicate universe $U$ such that an Attribute Clustering (AC) blocking scheme, as presented in \cite{hhis}, can be expressed as a single attribute-aware DNF scheme (Definition \ref{adnf}) that is a disjunction of all the predicates in $U$.
\end{theorem}

This theorem shows that, given a particular `reasonable' predicate universe, AC does not take into account node attribution (and is hence expressible as a single attribute-aware DNF scheme). On this account, a general (i.e. composite) DNF expression is strictly more expressive.

A second issue is that AC schemes are non-adaptive, and cannot be learned from training data (whether manually or automatically constructed). This implies that its performance may not be as competitive for `peculiar' datasets and domains. On the other hand, DNF schemes, in the formulation presented in this paper, can be learned using approximation techniques from the complexity-theory literature.
Finally, unlike AC, which requires access to the \emph{entire} dataset to formulate its predicates \cite{hhis}, DNF schemes only need access to limited training data. This gives them an advantage of scale in cases where the entire dataset, but not the required fraction of training examples, is too large to fit in memory.

\section{Conclusion}\label{conclusion}
In this paper, we presented a graph-theoretic construction for DNF schemes, applicable to a directed, labeled attributed data graph model. The presented schemes are functions that are useful for reducing pairwise (i.e. quadratic) complexity in sparse supervised machine learning-based edge-discovery on either a single data graph or between two data graphs. Previously, the DNF schemes had only been proposed for homogeneous tabular deduplication. Table \ref{contributions} summarizes the technical contributions in this work. An optimization-based framework can be used for learning the schemes. The empirical promise of these schemes (in terms of high volume, dynamicity and stability) was demonstrated in real-world settings against the competitive Attribute Clustering baseline.

{\bf Future Work.} Given the general applicability of DNF schemes, there are several (theoretical and practical) avenues for future work. One aspect that we are looking to investigate is to approximate a good DNF blocking scheme when the link specification function (e.g., friendship links between nodes) or LSF is \emph{known}. Thus far in this paper, we only covered the adaptive case when the LSF itself is unknown and will likely be approximated through an independent feature extraction and supervised machine learning pipeline (using the same training data as the DNF learner). In the general case, this problem is infeasible if the LSF is just treated as a black box. Some work has attempted solutions when the LSF is in a metric space e.g., Locality Sensitive Hashing has been used for complexity reduction with respect to LSFs such as Jaccard similarity. However, the problem is still relatively under-studied for non-metric LSFs. 

On an empirical front, we are also looking to expand beyond the case study in Section \ref{applications} and implement the principles in this paper for blocking large datasets in \emph{unusual} domains e.g., human trafficking and securities fraud. In recently processed datasets in these domains, we found severe entity resolution issues. Because each dataset contains many millions of nodes, scalable candidate generation is an important concern, for which we are looking to apply some of the adaptive methods described in this work.

{\bf Acknowledgements.} The author conducted this work in 2016 in the final semester of his PhD at the University of Texas at Austin. He gratefully acknowledges the support of his advisor, Daniel P. Miranker, and the fruitful discussions that ultimately led to the writing of the paper.  




\bibliographystyle{ACM-Reference-Format}
\bibliography{sigproc} 

\end{document}